%% file: main.tex
\documentclass[acmtog,nonacm]{acmart}

\def\ie{\emph{i.e., }}

\usepackage{multirow}
\usepackage{graphicx}

\usepackage{appendix}

\usepackage[ruled]{algorithm2e} 

\SetAlFnt{\small}
\SetAlCapFnt{\small}
\SetAlCapNameFnt{\small}
\SetAlCapHSkip{0pt}

\acmJournal{TOG}
\copyrightyear{2025}
\acmYear{2019}

\begin{document}
\title{DAM-VSR: Disentanglement of Appearance and Motion for Video Super-Resolution}

\author{Zhe Kong}
\affiliation{
\institution{Shenzhen Campus of Sun Yat-sen University, China, Meituan, China and Division of AMC and Department of ECE, HKUST}
\country{Hong Kong}}


\author{Le Li}
\affiliation{
\institution{Tianjin University}
\country{China}}

\author{Yong Zhang}
\authornote{Corresponding authors. \\ 
Project~: ~\url{https://kongzhecn.github.io/projects/dam-vsr/}}
\author{Feng Gao}
\affiliation{
\institution{Meituan}
\country{China}}

\author{Shaoshu Yang}
\affiliation{
\institution{School of Artificial Intelligence, University of Chinese Academy of Sciences}
\country{China}}

\author{Tao Wang}
\affiliation{
\institution{Nanjing University}
\country{China}}

\author{Kaihao Zhang}
\affiliation{
\institution{Harbin Institute of Technology, Shenzhen}
\country{China}}

\author{Zhuoliang Kang}
\author{Xiaoming Wei}
\affiliation{
\institution{Meituan}
\country{China}}

\author{Guanying Chen}
\affiliation{
\institution{Shenzhen Campus of Sun Yat-sen University}
\country{China}}

\author{Wenhan Luo}
\authornotemark[1]
\affiliation{
\institution{Division of AMC and Department of ECE, HKUST}
\country{Hong Kong}}

\begin{abstract}
Real-world video super-resolution (VSR) presents significant challenges due to complex and unpredictable degradations. Although some recent methods utilize image diffusion models for VSR and have shown improved detail generation capabilities, they still struggle to produce temporally consistent frames. We attempt to use Stable Video Diffusion (SVD) combined with ControlNet to address this issue. However, due to the intrinsic image-animation characteristics of SVD, it is challenging to generate fine details using only low-quality videos. To tackle this problem, we propose DAM-VSR, an appearance and motion disentanglement framework for VSR. This framework disentangles VSR into appearance enhancement and motion control problems. Specifically, appearance enhancement is achieved through reference image super-resolution, while motion control is achieved through video ControlNet. This disentanglement fully leverages the generative prior of video diffusion models and the detail generation capabilities of image super-resolution models. Furthermore, equipped with the proposed motion-aligned bidirectional sampling strategy, DAM-VSR can conduct VSR on longer input videos. DAM-VSR achieves state-of-the-art performance on real-world data and AIGC data, demonstrating its powerful detail generation capabilities.
\end{abstract}


\begin{teaserfigure}
    \centering
  \includegraphics[width=1.0\textwidth]{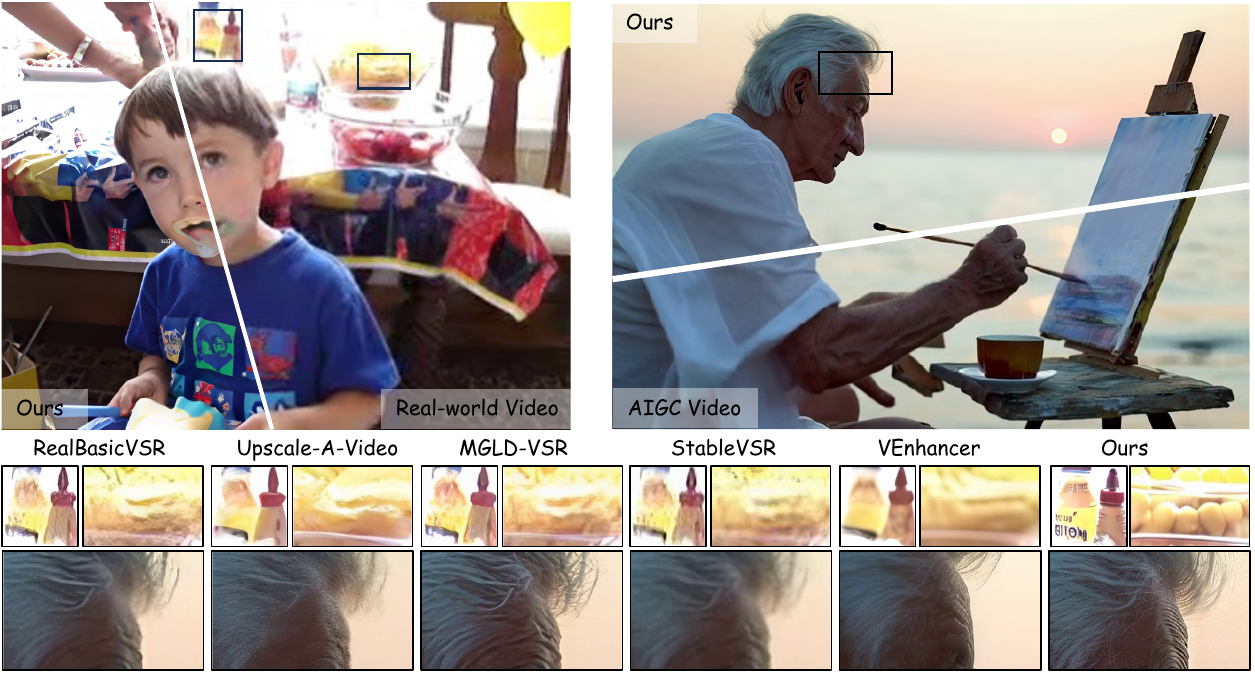}
  \caption{Visualization comparisons with state-of-the-art methods on both real-world and AIGC videos. Our DAM-VSR demonstrates remarkable upscaling capabilities, generating more realistic details with better visual realism than other methods. (\textbf{Zoom-in for the best view})}
  \label{fig:teaser}
\end{teaserfigure}

\maketitle

\input{samplebody-journals}

\end{document}


\title{Supplementary Materials}

\maketitle

\noindent This appendix includes our supplementary materials related to the proposed DAM-VSR. The structure of the supplementary materials is as follows:
\begin{itemize}
    \item \textcolor{black}{Details of tile sampling for higher resolution video generation (Section~\ref{sec:tile-sampling})}
    \item \textcolor{black}{Details of our training strategy (Section~\ref{sec:training})}
    \item \textcolor{black}{Details of image super-resolution (Section~\ref{sec:isr})}
\end{itemize}

\section{Details of tile sampling for higher resolution video generation}
\label{sec:tile-sampling}

Video diffusion model requires substantial memory during the inference process, which poses challenges for directly generating high-resolution videos with limited memory resources. 
Recently, high-resolution image synthesis has achieved remarkable success, with numerous methods proposing techniques to generate images at much higher resolutions than the training image sizes through overlapped patch-based denoising. 
Inspired by these advancements, we utilize a tile diffusion sampling method for VSR to enhance the proposed method's capability to produce higher-resolution outputs with limited memory.

Specifically, as illustrated in Fig. \ref{fig:spatial-upscale}, we begin by splitting the input videos into a series of overlapping blocks. Subsequently, the conventional denoising process is applied independently to each block. Afterward, the overlapping regions of adjacent blocks are averaged in the latent space after each sample step. 
After denoising, the VAE decoding process is conducted in the same manner as the diffusion process, which involves decoding each block to obtain the RGB video block separately. Finally, we blend these RGB blocks to produce a high-resolution output video. With the tile sample, the VSR can overcome the limited memory constraint to conduct video super-resolution on higher-resolution input videos.

\setcounter{figure}{10}
\begin{figure}[htb]
  \centering
  \includegraphics[width=0.98\linewidth]{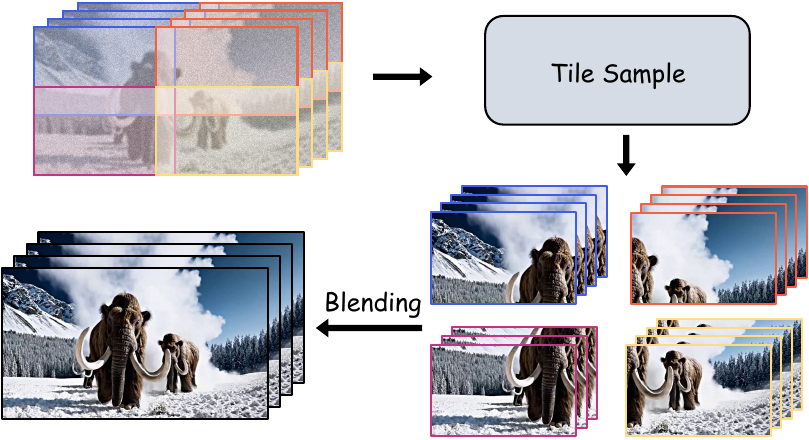}
  \caption{Illustration of tile sampling, which can handle the generation of high-resolution inputs.}
  \label{fig:spatial-upscale}
\end{figure}

\section{Details of our training strategy}
\label{sec:training}

The training process is divided into three stages. 

In the first stage, we first train a video ControlNet conditioned on a low-quality video clip and a high-quality reference image. The overall network architecture during this stage is illustrated in Fig. 4. The architecture consists of a denoising UNet and a video ControlNet. Only the weights of the video ControlNet are updated during training.

In the second stage, the entire training process consists of three networks: the forward UNet $\epsilon_\theta$, the video ControlNet $CT_\alpha$ derived in stage $1$, and the backward UNet $\epsilon_\theta^{'}$. During training, we first calculate the temporal attention map $A_t$ leveraging the forward UNet $\epsilon_\theta$ and video ControlNet $CT_\alpha$. 
Then, conditioned on the the rotated attention map $A_t^{'}$, reversed low-quality video $C^{'}$ and high-quality end frame $H_k$, we predict the noise $P_{t}^{b}$ of reverse motion utilizing the backward UNet $\epsilon_\theta^{'}$. And then calculating the training loss for updating.
We follow and only fine-tune the $W_v$ and $W_o$ in the temporal self-attention in $\epsilon_\theta^{'}$. Video ControlNet $CT_\alpha$ and forward UNet $\epsilon_\theta$ keep frozen during training.
The first and second stages utilize only diffusion loss for training. During inference, $CT_\alpha$ and $CT_\alpha^{'}$ share the same weight.

The third stage primarily focuses on fine-tuning the VAE decoder. We first generate the predicted high-quality videos through $8$ steps of sampling. Then fine-tune the VAE decoder using the corresponding ground-truth video sequence. During training, we employ $L2$ loss, perceptual loss, and GAN loss. The training objective can be summarized as follows:
\begin{equation}
\mathcal{L} =  \mathcal{L}_{L2} + \alpha \mathcal{L}_{percept} + \beta \mathcal{L}_{GAN},
\end{equation}
where $\alpha$ and $\beta$ are set to $1$ and $0.025$, respectively.

The training data is sampled from the dataset at a resolution of $576 \times 1024$ using random cropping, with a frame length of $14$. In stage 1, only the video ControlNet is trained for $60k$ iterations. During stage 2, we fine-tune only the $W_v$ and $W_o$ in temporal attention of the UNet for another $45k$ iterations. Finally, in the last stage, we fine-tune the decoder of the VAE using LoRA for another $10k$ iteration. Throughout the training process, we adhere to the degradation pipeline of to generate low-quality and high-quality video pairs.




\section{Details of image super-resolution}
\label{sec:isr}

The ISR methods, such as ResShift and InvSR, utilize latent consistency models, requiring only $1$ or $4$ steps, which significantly decrease the inference time. Additionally, unlike most methods requiring $50$ sampling steps, our SVD-based approach requires only $30$. Using SDEdit, we add noise equivalent to 60\% of the steps to the low-quality input as initial noise, effectively reducing the number of sampling steps to $18$, minimizing inference time.



%% file: samplebody-journals.tex
\section{Introduction}
Video super-resolution (VSR) aims to generate high-resolution videos that are motion-consistent with the input low-resolution videos.
Most previous VSR methods \cite{chan2021basicvsr,liang2024vrt,wang2024gridformer,tan2023blind} focus on either synthetic degradations or specific camera-related degradations. Since real-world low-resolution videos often exhibit more complex and unpredictable degradations, it is challenging to generalize well to such videos.

Some GAN-based methods \cite{chan2022investigating, xie2023mitigating} utilize generative priors to solve real-world VSR. Although they have demonstrated potential success in generating clear video structures with smooth motion across frames, they still fall short in producing fine details with realistic textures due to limited generative capabilities, as shown in Fig. \ref{fig:teaser}. Recently, diffusion models \cite{ho2020denoising, rombach2022high} have shown remarkable performance in generating images and videos with realistic, fine-grained details. Some VSR methods \cite{zhou2024upscale, yang2025motion, rota2023enhancing, tan2024blind} adopt image diffusion models to perform frame-by-frame super-resolution in the latent space. These methods employ optical flow warping or introduce additional temporal layers to improve motion consistency. Despite these improvements, they still produce inconsistent details across consecutive frames, which disrupts the visual coherence of the generated video.

Hence, several methods propose to utilize video diffusion models to improve temporal consistency. For instance, VEnhancer \cite{he2024venhancer} employs a text-to-video diffusion model for VSR on low frame rate and low-resolution videos. However, it struggles to maintain fidelity, rendering it unsuitable for real-world VSR. SeedVR \cite{wang2025seedvr} proposes a shifted window attention mechanism that facilitates effective restoration on long video sequences.  It extends the image diffusion model SD3 \cite{esser2024scaling} to a video diffusion model and treats the VSR task as a video diffusion model training problem. This training process utilizes approximately $100$ million images and $5$ million videos, trained on 256 NVIDIA H100-80G GPUs, representing excessive data and GPU consumption.

\begin{figure}[t]
  \centering
  \includegraphics[width=0.98\linewidth]{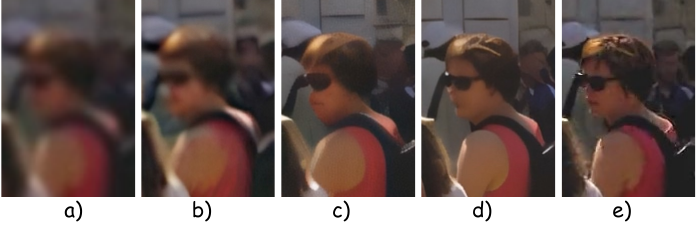}
  \caption{Analysis of the role of ISR enhancement. a) Input video. b) The generated result without ISR enhancement. c) The generated video with ISR enhancement. d) The generated result of our method. e) Ground truth.}
  \label{fig:motivation}
\end{figure}

To fully leverage the generative prior of the video model, we propose using the Stable Video Diffusion (SVD) model for VSR to alleviate the demand for extensive training data. In image super-resolution (ISR), numerous methods \cite{yu2024scaling,lin2025diffbir,chen2024towards} utilize the powerful structural preservation capabilities of ControlNet to achieve enhanced results. Following these approaches, we initiate VSR by utilizing video ControlNet, where the low-quality video is input for motion and structure control. The video's details are then supplemented with the generative capabilities of the video diffusion model, thereby completing the super-resolution process. However, as shown in Fig. \ref{fig:motivation}, our attempts did not yield satisfactory results. Although there is a slight improvement, the effect is not significant, as shown in Fig. \ref{fig:motivation}b.
We argue that SVD's reference image animation characteristics cause this. SVD is an image-to-video diffusion model that can animate the provided reference image to generate a video. The overall appearance of the generated video is controlled by the reference image. However, due to the lack of fine details in the reference image, relying solely on video ControlNet cannot effectively enhance the appearance of VSR. 
Recent advancements in ISR methods \cite{yu2024scaling,yue2024resshift,yue2024arbitrary} have made significant progress, enabling the generation of input images with more realistic textures. The development of ISR is ahead of that of VSR, and we believe that VSR can benefit from ISR. Therefore, an intuitive approach is to use ISR to assist VSR by providing visual enhancements. Hence, we make such an attempt and retrain a ControlNet conditioned on low-quality videos and high-quality reference images. The results are shown in Fig. \ref{fig:motivation}c, and performance is improved, verifying our analysis.

Based on these analyses, the overall framework is established. We propose DAM-VSR, a novel appearance and motion disentanglement framework to generate high-resolution videos with realistic details and temporal-consistent frames.
We disentangle the VSR challenge into two distinct tasks: appearance enhancement and motion control.
Appearance enhancement is addressed through reference image ISR.
We first perform ISR on the reference image. Once the reference image is enhanced, the realistic details in the reference frame can be propagated to the remaining frames of the generated video through the I2V model, enabling us to produce a high-resolution video with temporal consistency. Besides, by selecting different ISR methods, this approach allows for the generation of videos with a trade-off between high fidelity and better perceptual quality.
On the other hand, low-resolution input videos provide motion control through ControlNet. 
Unlike SeedVR \cite{wang2025seedvr}, our method benefits from the appearance enhancement provided by the ISR method and temporal consistent by SVD, eliminating the reliance on extensive training data.
Despite the effectiveness of the proposed disentanglement framework, the frame length of the generated videos is constrained to a specific number of frames due to the limitations of the base I2V model. This constraint complicates the application of VSR for long video generation. To tackle this issue, we propose a motion-aligned bidirectional sampling strategy for long video super-resolution.  This sampling strategy consists of a forward and a backward generation while ensuring temporal consistency through motion alignment, facilitating the long video generation.

In summary, our main contributions are as follows.
\begin{itemize}
\item We propose DAM-VSR, an appearance and motion disentanglement framework for real-world VSR that fully leverages the generative prior of video diffusion models and the detail generation capabilities of ISR models. By choosing different ISR methods, the generated videos achieve a trade-off between high fidelity and enhanced perceptual quality.
\item We propose a motion-aligned bidirectional sampling strategy for long VSR, which effectively alleviates the flickering issue encountered in long video generation.
\item Our DAM-VSR has achieved state-of-the-art performance on real-world data and AIGC data, demonstrating the effectiveness of our proposed method and showcasing remarkable perceptual quality and temporal consistency.
\end{itemize}

\section{Related Work}
\label{sec:related}

\subsection{Video Super-Resolution}
Video super-resolution methods aim to enhance low-resolution video frames to high-resolution ones. Traditional methods \cite{chan2021basicvsr,chan2022basicvsr++,isobe2020video,liang2022recurrent,xue2019video} are usually based on stacked CNN or transformer layers and are trained under simple degradation (e.g., bicubic resizing, downsampling, blur, and noise). Since real-world low-resolution videos often exhibit more complex and unknown degradations, models trained with such synthetic data frequently experience a noticeable performance drop when applied to real-world low-resolution data. To bridge this gap, some methods \cite{chan2022investigating,xie2023mitigating,xu2024videogigagan} synthesize more realistic training data by mixing diverse types of degradations. Although these methods have achieved better performance, due to the lack of generative priors, they still cannot generate photo-realistic details and textures.

Recently, image diffusion models have demonstrated remarkable progress in generating realistic and diverse images from text prompts \cite{rombach2022high,li2024photomaker,kong2025omg}. Some VSR methods \cite{zhou2024upscale,yang2025motion,rota2023enhancing,tang2024seeclear} seek to derive more generative priors from these text-to-image models. For example, Upscale-A-Video \cite{zhou2024upscale} utilizes a pre-trained image $x4$ upscaling model for super-resolution. Additionally, a training-free flow-guided recurrent latent propagation module is employed to ensure temporal consistency. StableVSR \cite{rota2023enhancing} also utilizes an image diffusion model as its base model and designs a temporal texture guidance method to generate temporally consistent frames. MGLD-VSR \cite{yang2025motion} proposes a motion-guided loss that generates temporally consistent latent features. A temporal-aware sequence decoder is proposed to further enhance the continuity of the generated videos. SeeClear \cite{tang2024seeclear} proposes an instance-centric alignment module to enhance frame coherency and a channel-wise texture aggregation memory to fully leverage semantic textures. Despite significant improvements, these methods all utilize the image diffusion model and do not obtain priors from video diffusion models, resulting in temporal inconsistencies in the latent space. 

Later, some methods that utilize the video diffusion model for VSR have also been proposed. VEnhance \cite{he2024venhancer} is a video enhancement method based on a video diffusion model that supports both appearance and temporal super-resolution. STAR \cite{xie2025star} introduces a local information enhancement module to enrich local details and mitigate degradation artifacts. It also proposes a dynamic frequency loss to reinforce fidelity. SeedVR \cite{wang2025seedvr} proposes a shifted window attention mechanism for long video sequences. Additionally, it extends SD3 to a video diffusion model and utilizes a causal video autoencoder, mixed image and video training, and progressive training for video diffusion training.

\begin{figure*}[t]
  \centering
  \includegraphics[width=0.90\linewidth]{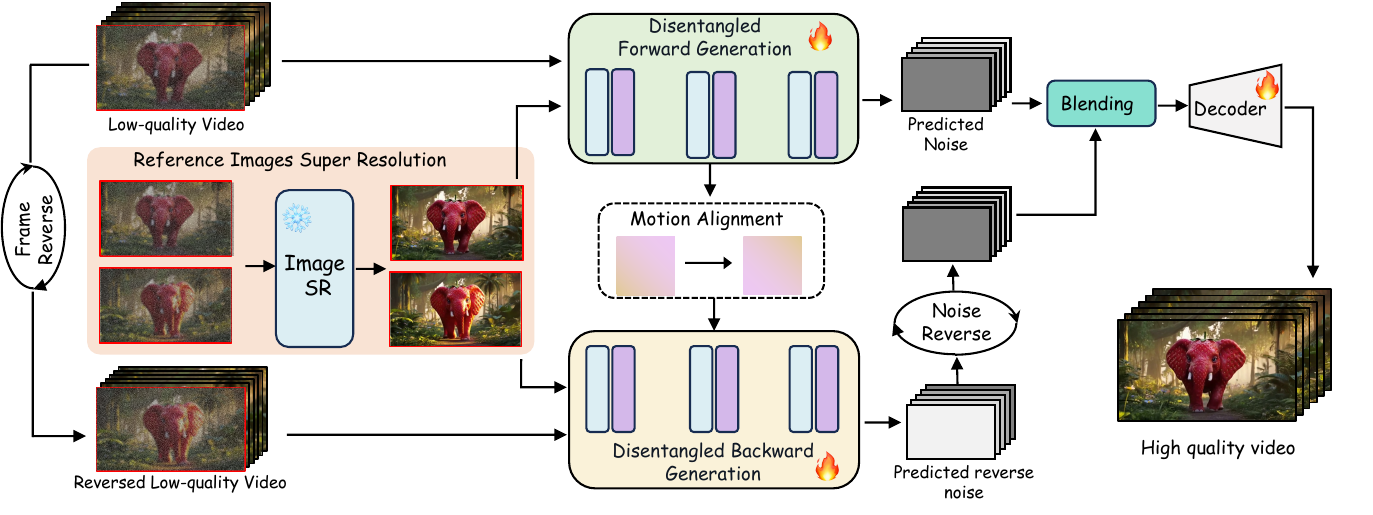}
  \caption{The overall pipeline of the proposed DAM-VSR method. Our method introduces an appearance and motion disentanglement framework for VSR. To support long video generation, we propose a motion-aligned bidirectional sampling strategy, which consists of a disentangled forward generation process and a disentangled backward generation process. These two processes maintain temporal consistency through motion alignment.
   }
  \label{fig:pipeline}
\end{figure*}

\subsection{Video Diffusion Models}
In recent years, the success of diffusion models in text-to-image generation has sparked considerable interest in exploring their application for video generation. Video generation models \cite{chen2023videocrafter1, chen2024videocrafter2, xing2025dynamicrafter, wang2023modelscope, blattmann2023stable, yang2024cogvideox, kong2024hunyuanvideo, brooks2024video} have made remarkable progress, attributed to the strong generative capabilities of model architecture and the abundance of training data. These models can be broadly categorized into two types: text-to-video generation models \cite{chen2023videocrafter1, chen2024videocrafter2, wang2023modelscope} and image-to-video generation models \cite{xing2025dynamicrafter, blattmann2023stable}.
LDV \cite{ho2022video} is the first attempt to apply diffusion models to video generation. Trained on a well-curated video dataset, SVD \cite{blattmann2023stable} proposes a unified strategy for developing a robust video generation model. Early video diffusion models typically leverage the U-Net architecture for video generation. More recently, diffusion transformer-based video diffusion models \cite{ma2024latte}, such as CogVideoX \cite{yang2024cogvideox}, Sora \cite{brooks2024video}, and Hunyuan \cite{kong2024hunyuanvideo}, have demonstrated enhanced video generation capabilities. These video diffusion models have provided a strong visual backbone for various downstream tasks \cite{zhao2024stereocrafter}, including image animation \cite{hu2024animate, xue2024follow}, video style transfer \cite{liu2024stylecrafter, ye2024stylemaster}, video editing \cite{qi2023fatezero}, and frame interpolation \cite{wang2024generative,yang2024zerosmooth}.

\section{Method}


The overall architecture of our proposed method is illustrated in Fig. \ref{fig:pipeline}, which presents an appearance and motion disentanglement framework for VSR. To support VSR for long videos, we introduce a motion-aligned bidirectional sampling strategy that incorporates both a forward generation and a backward generation to maintain temporal consistency at each timestep.
In the following, we first review the video diffusion model in Sec. \ref{pre}. Next, we introduce our appearance and motion disentanglement framework in Sec. \ref{decomposition-framework}. In Sec. \ref{bidrectional-sample}, we present our motion bidirectional sampling strategy for long video generation. Finally, we discuss our disentanglement framework to other applications in Sec. \ref{applications}.


\subsection{Preliminary}
\label{pre}

Latent diffusion model \cite{ho2020denoising,rombach2022high} belongs to a class of generative models containing a forward process and a reverse process in the latent space, which is trained to convert random noise into high-resolution images/videos through an iterative sampling process. Stable Video Diffusion (SVD) \cite{blattmann2023stable} is an image-to-video diffusion model condition on a reference image. In the forward process, a video clip $x$ is firstly projected to latent space through VAE encoder $z_0 = \mathcal{E}(x)$. Then a random Gaussian noise $\epsilon$ at time step $t$ is added to the data sample $z_0$ as $z_t = \alpha_t z_0 + \sigma_t \epsilon$, where $\alpha_t$ and $\sigma_t$ is predefined noise adding schedule. The key objective of the diffusion model is to train a denoising U-Net $\epsilon_\theta$ conditioned on the reference image $c$ defined as
\begin{equation}
\mathcal{L(\theta)} = E_{t \sim U[1, T]}||\epsilon_\theta(z_t, t, c) - y||^2_2,
\end{equation}
where $y=\alpha_t \epsilon - \sigma_t z_t$, referred to as v-prediction.

\subsection{Appearance Motion Disentanglement Generation}
\label{decomposition-framework}

\begin{figure}[t]
  \centering
  \includegraphics[width=0.98\linewidth]{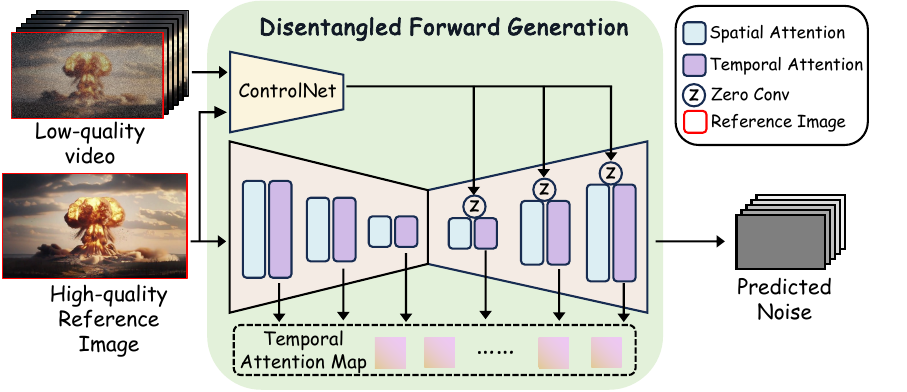}
  \caption{The model architecture of the disentangled forward generation (the same network architecture as the disentangled backward generation) includes a video ControlNet and a denoising UNet. The input for the generation consists of a low-quality video and a high-quality reference image.}
  \label{fig:decomp-generation}
\end{figure}

Given a low-quality video clip $C=\{I_1, I_2, \cdots, I_k\}$ composed of $k$ frames (the same length as SVD, $14$ in our case), our goal is to generate the corresponding high-quality video clip leveraging the generative prior of SVD.

SVD is an image-to-video diffusion model conditioned on a reference image, capable of animating a reference image (first frame) to generate a video. We adopt SVD as our base model to fully leverage the generative prior for VSR. 
For SVD, the reference image provides the appearance of the generated video, which determines the overall appearance of the generated video.
We find that high-quality reference images, characterized by detailed textures, significantly enhance the richness of the generated video, resulting in a visually compelling output. Conversely, low-quality reference images detract from the appeal of the generated video. Based on this, we propose an appearance and motion disentanglement framework for VSR. We disentangle VSR tasks into two challenges: appearance enhancement and motion control. Appearance is provided by the reference image, while motion is controlled by the low-quality video.


Appearance enhancement is addressed through reference image super-resolution. Recently, ISR has made significant progress. Numerous ISR methods \cite{yu2024scaling, yue2024resshift, yue2024arbitrary} have been proposed and can generate realistic textures and clear details from low-quality images. Some of these methods focus on fidelity, while others emphasize generation quality. By choosing different ISR methods, it is possible to achieve either high fidelity or enhanced perceptual quality in generating images.
A promising approach to achieve appearance enhancement involves conducting ISR on the reference frame of the input video clip, represented by:
\begin{equation}
H_1 = SR_I(I_1),
\end{equation}
where $SR_I$ denotes the ISR model, $I_1$ is the first frame of $C$ and $H_1$ is the generated high-quality first frame with realistic details. 

Subsequently, these generated realistic details in the reference frame can be propagated to the remaining frames leveraging the powerful temporal generation capabilities of SVD. This process ensures that a temporally consistent video is produced.
On the other hand, VSR needs to ensure that the motion of the generated video is consistent with that of the original video. Therefore, to achieve motion control, we utilize video ControlNet, which takes the input video clip $C$ as a condition and controls the overall temporal dynamics of the generated video. 
As shown in Fig. \ref{fig:decomp-generation}, our disentanglement generation framework takes the low-quality video $C$ and the high-quality reference image $H_1$ as inputs and predicts the corresponding noise in each step. 
Specifically, the high-quality reference image $H_1$ is first encoded by VAE. Then, after $k$ times repeating and then the image latent concatenates with a random noise in the channel dimension. This concatenated latent serves as an initial latent for UNet and video ControlNet. Additionally, $H_1$ is injected into UNet and video ControlNet through cross attention after the clip encoder. For $C$, it provides motion control for SVD through input to the video ControlNet after several video embedding layers.

For video ControlNet, we adapt the design from image ControlNet \cite{zhang2023adding}. Firstly, we duplicate the original UNet’s encoder, cloning all the layers, including temporal attention and 3D convolutional layers, along with their weights. 
Secondly, we add several video embedding layers to encode the low-quality videos and add them with the concatenated feature of latent noise and reference image. Finally, the outputs of each layer of the video ControlNet are added to the skip connections of the original U-Net decoder.

Although VSR can be achieved through the proposed disentanglement framework, video diffusion models require substantial memory during the inference process, which poses challenges for directly generating higher-resolution videos with limited memory resources. To enable our method to handle high-resolution input videos, we utilize tile sampling \cite{bar2023multidiffusion, du2024demofusion, he2023scalecrafter} to alleviate excessive memory consumption. More details can be found in the appendix.

\subsection{Long Video Super-Resolution}
\label{bidrectional-sample}

Despite the substantial generative capabilities offered by the SVD, the frame length of the generated videos is severely constrained, and limited to a specific number of frames. The computational complexity scales quadratically with the number of frames, which presents significant challenges for the direct generation of long videos. To solve this problem, a straightforward approach involves dividing the input video into several clips and performing VSR on each clip individually. Finally, the processed clips can be concatenated to produce the final result. However, this can lead to flickering between clips, making it increasingly challenging to maintain consistency throughout the video.
An effective strategy to mitigate flickering is to adopt an auto-regressive approach for generating the long video. Specifically, the last frame of each generated video clip is adopted as the reference image (first frame) for the subsequent clip. However, it results in error accumulation.

\begin{figure}[t]
  \centering
  \includegraphics[width=0.85\linewidth]{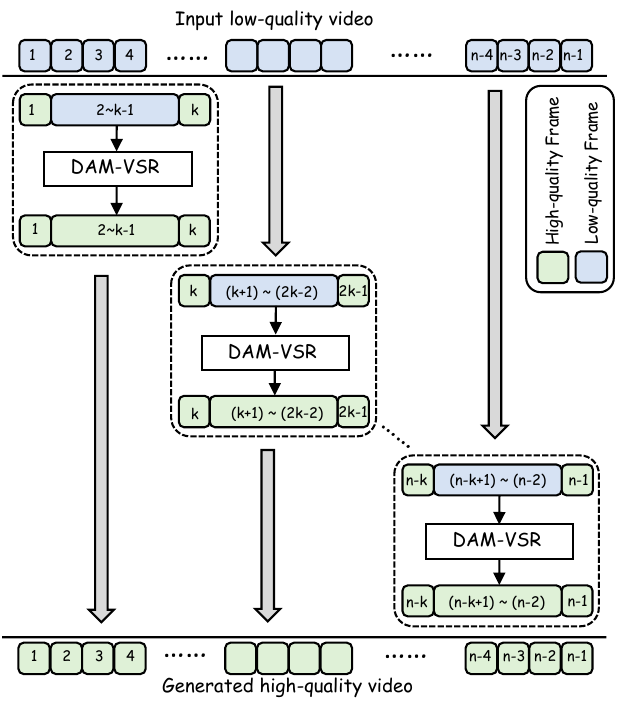}
  \caption{Illustration of long video processing. Through clip concatenation, our method supports VSR for long input videos.}
  \label{fig:temp-upscale}
\end{figure}

Hence, to address the video flicker issue and avoid error accumulation, we propose a motion-aligned bidirectional sampling method for long video super-resolution, as shown in Fig. \ref{fig:temp-upscale}. The core idea of this approach is to divide the input video into multiple clips with overlapping starting and ending frames between adjacent video clips. 
We first conduct ISR on the starting and ending frames of each clip to obtain the corresponding high-quality images. Then, for each clip, we generate high-quality intermediate frames using our proposed motion-aligned bidirectional sampling strategy, conditioned on the high-quality starting frame, ending frame, and low-quality video clip. Since different clips share the same high-quality starting and ending frames with adjacent video clips, this ensures that the final concatenated long video maintains motion consistency.


\subsubsection{Motion-aligned Bidirectional Sampling}

In motion-aligned bidirectional sampling, we aim to generate high-quality intermediate frames conditioned on the high-quality starting frame, the ending frame, and the low-quality video clip.
During sampling, each step involves two processes: a disentangled forward generation and a disentangled backward generation. The two sample processes utilize motion alignment to maintain temporal consistency, as shown in Fig. \ref{fig:pipeline}.
Given a low-quality video clip $C=\{I_1, I_2, \cdots, I_k\}$, we firstly conduct ISR on the starting frame $I_1$ and ending frame $I_k$ to obtain the corresponding high-quality frame $H_1$ and $H_k$.
And then start our motion-aligned bidirectional sampling.
Specifically, the disentangled forward generation is mainly for predicting the forward noise for the generated video. At timestep $t$, the forward sample process can be represented as:
\begin{equation}
p_t^{f}, A_t = F(z_t, t, H_1, C),
\end{equation}
where $z_t$ is the latent noise, $p_t^{f}$ and $A_t$ represent the predicted noise and temporal attention map in the forward process. The network architecture during forward generation $F$ is shown in Fig. \ref{fig:decomp-generation}, which consists of a denoising UNet $\epsilon_\theta$ and a video ControlNet $CT_\alpha$.

On the contrary, the disentangled backward generation is exactly the opposite process, which involves a reverse noise prediction process. During the backward sample, the generation is conditioned on the end frame $H_k$ and the reverse low-quality clip $C^{'}$. 
Since the temporal self-attention maps $A_t$ can reflect the motion dynamics of $C$ \cite{wang2024generative}, to obtain the backward motion of $C$, we align the motion by rotating the attention maps of $A_t$ by $180$ degrees to derive $A_t^{'}$. The modified attention map $A_t^{'}$ reflects the motion of $C^{'}$. 
Specifically, the $x$-th row and $y$-th column in $A_t$ reflect the relation of $x$-th and $y$-th frame in $C$. In the reversed video clip, $A_{k-x, k-y}^{'}=A_{x,y}$. And the $(k-x)$ frame and $(k-y)$ frame still maintain the same relation.
Hence, the backward process shares the rotated temporal self-attention maps with the forward process to synchronize the sample path. 
Besides, to maintain motion consistency, the backward generation shares the same latent noise with the forward generation but with a reverse frame. The reverse noise $z_t^{'}$ can be derived through frame reverse, denoted as $z_t^{'} = reverse(z_t)$.
This generation process is achieved by replacing the generated temporal attention map $A_t^{g}$ with $A_t^{'}$ in the sampling process:
\begin{equation}
\label{eq:align}
p_t^{b}, \_ =B(z_t^{'}, t, H_{k}, C_{'})\{A_t^{g} \leftarrow A_t^{'}\},
\end{equation}
where $p_t^{b}$ denotes the predicted backward noise.
The backward generation shares the same network architecture as the forward process, but there are slight differences in network weights. The video ControlNet $CT_\alpha^{'}$ in $B$ shares the weights with $CT_\alpha$. The UNet $\epsilon_\theta^{'}$ in $B$ shares most of its weight with $\epsilon_\theta$, where the only difference is the weights associated with the temporal self-attention layer.

After backward generation, we reverse the predicted noise $p_t^{b}$ to obtain the correct frame order and subsequently blend it with the forward noise $p_t^{f}$. This process ultimately yields the final predicted noise $p_t$ at timestep $t$ through
\begin{equation}
p_t = blending(reverse(p_t^{b}), p_t^{f}),
\end{equation}
where for blending, we employ the simplest averaging operation.

Through updating, we can get $z_{t-1}$ can from $z_t$ and $p_t$. The motion-aligned bidirectional sampling performs at each sampling step until $t=0$. Then, after VAE decoding, we obtain a high-resolution video.

\subsubsection{Long Video Generation through Clip Concatenating}

Combined with motion-aligned bidirectional sampling, the proposed method supports long video generation through clip concatenating.
Specifically, the process is shown in Fig. \ref{fig:temp-upscale}, given a long low-quality video $I=\{I_1, I_2, \cdots, I_n\}$ composed of $n$ frames, where $n>>k$. The input $I$ is initially divided into $m$ video clips, defined as follows:
\begin{equation}
I=\{I_1, I_2, \cdots, I_n\}=\{C_1, C_2, \cdots, C_i, \cdots, C_m\}.
\end{equation}
Each video clip is of uniform length and comprises $k$ frames (for simplicity, we ignore the last remaining frames), expressed as:
\begin{equation}
\begin{split}
&C_1 = \{I_1, I_{2}, \cdots, I_{k}\}, \\ 
&C_i = \{I_j, I_{j+1}, \cdots, I_{j+k-1}\}, \\ 
&C_m = \{I_{n-k+1}, I_{n-k+2}, \cdots, I_{n}\}.
\end{split}
\end{equation}


For any video clip $C_i$ ($1<i<m$), we first perform ISR on the starting frame $I_j$ and the ending frame $I_{j+k-1}$. Then, conditioned on $C_i$ and the generated high-quality staring and ending frames $H_j$ and $H_{j+k-1}$, we can obtain the corresponding high-quality video clip $\{H_{j} \sim H_{j+k-1}\}$ through motion-aligned bidirectional sampling.
For next time, we performance VSR on $C_{i+1}=\{I_{j+k-1} \sim I_{j+2k-2}\}$. Since $C_i$ and $C_{i+1}$ share the same high-quality frame $H_{j+k-1}$, this effectively alleviates flickering during video clip concatenation. Ultimately, it achieves long video generation for VSR.

\subsection{Other Applications}
\label{applications}

In addition to VSR, the appearance and motion disentanglement framework proposed in this paper can be applied to various video-related applications, including video editing and video style transfer. Due to the unsuitability of RGB-controlled video ControlNet for image editing and style transfer, we additionally trained a video ControlNet conditioned on video depth and utilized this depth-conditioned ControlNet for other applications. For video editing, we first edit the reference image utilizing Ledits++\cite{brack2024ledits++} and then generate the edited video conditioned on the edited reference and the original video. The experimental results are shown in the upper part of Figure \ref{fig:application}, where we successfully changed the swan's feathers from black to blue. For video style transfer, we follow the same pipeline that modifies the reference image through InstantStyle \cite{wang2024instantstyle} and then generates the transferred video. The visual result is shown in the lower part of Figure \ref{fig:application}.



\section{Experiments}

\begin{table}[]
\centering
\caption{Ablation study of various components within DAM-VSR on UDM10.}
\label{tab:ablation}
\resizebox{.90\columnwidth}{!}{%
\begin{tabular}{l|c|c|c}
\hline
\multicolumn{1}{c|}{}                      & PSNR $\uparrow$ & SSIM $\uparrow$& LPIPS $\downarrow$\\ \hline
a) Baseline                                   & 24.775          & 0.739          & 0.376          \\
b) Frame-by-frame SR                          & 25.561          & 0.713          & 0.393          \\
c) w/o Disentanglement During Training and Inference & 25.379          & 0.752          & 0.372          \\
d) w/o Disentanglement Only During Training   & 24.691          & 0.744          & 0.328          \\
e) w/o FT-VAE-Decoder                         & 26.851          & 0.751          & 0.382          \\
f) w/o Bidirectional Sample                   & 26.654          & 0.767          & 0.319          \\ \hline
g) Ours                                       & \textbf{27.011} & \textbf{0.776} & \textbf{0.311} \\ \hline
\end{tabular}%
}
\end{table}

\begin{table*}[t]
\centering
\caption{Quantitative evaluations on different VSR benchmarks from diverse source, \ie synthetic (UDM10, YouHQ40, REDS30), real-world (VideoLQ), and AIGC (AIGC29) data. The best and second best performances are marked in \textbf{\textcolor{red}{red}} and \textcolor{blue}{blue}, respectively.}
\label{tab:compare}
\setlength\tabcolsep{20pt}
\resizebox{1.90\columnwidth}{!}{%
\begin{tabular}{c|c|c|c|c|c|c|c}
\hline
Datasets                  & Metric   & RealBasicVSR & Upscale-A-Video & StableVSR & MGLD-VSR & VEnhancer & Ours \\ \hline
                          & PSNR $\uparrow$ & 25.852 & 24.181 & 25.989 & \textcolor{blue}{26.121} & 16.126 & \textcolor{red}{\textbf{27.011}} \\
                          & SSIM $\uparrow$ & 0.752 & 0.667 & \textcolor{blue}{0.755} & \textcolor{blue}{0.755} & 0.585 & \textcolor{red}{\textbf{0.776}} \\
                          & LPIPS $\downarrow$ & 0.322 & 0.369 & 0.389 & \textcolor{red}{\textbf{0.292}} & 0.488 & \textcolor{blue}{0.311} \\
\multirow{-4}{*}{UDM10}   & $E_{warp}$ $\downarrow$ & 1.504 & 2.125 & \textcolor{blue}{1.314} & 1.780 & \textcolor{red}{\textbf{1.060}} & 1.322 \\ \hline
                          & PSNR $\uparrow$ & 23.521 & 22.194 & 23.570 & \textcolor{blue}{23.920} & 16.301 & \textcolor{red}{\textbf{24.246}} \\
                          & SSIM $\uparrow$ & 0.633 & 0.546 & 0.597 & \textcolor{blue}{0.642} & 0.493 & \textcolor{red}{\textbf{0.668}} \\
                          & LPIPS $\downarrow$ & 0.390 & 0.422 & 0.473 & \textcolor{blue}{0.371} & 0.509 & \textcolor{red}{\textbf{0.367}} \\
\multirow{-4}{*}{YouHQ40} & $E_{warp}$ $\downarrow$ & 2.113 & 3.016 & 2.983 & 2.195 & \textcolor{red}{\textbf{1.205}} & \textcolor{blue}{1.769} \\ \hline
                          & PSNR $\uparrow$ & 23.517 & 21.388 & 23.474 & \textcolor{red}{\textbf{23.543}} & 12.433 & \textcolor{blue}{23.525} \\
                          & SSIM $\uparrow$ & \textcolor{red}{\textbf{0.613}} & 0.498 & 0.581 & 0.609 & 0.384 & \textcolor{blue}{0.612} \\
                          & LPIPS $\downarrow$ & \textcolor{blue}{0.338} & 0.443 & 0.504 & \textcolor{red}{\textbf{0.298}} & 0.648 & 0.382 \\
\multirow{-4}{*}{REDS30}  & $E_{warp}$ $\downarrow$ & 3.173 & 5.173 & 2.975 & 3.995 & \textcolor{red}{\textbf{2.342}} & \textcolor{blue}{2.927} \\ \hline \hline
                          & MUSIQ $\uparrow$ & \textcolor{blue}{55.032} & 49.260 & 25.155 & 50.943 & 49.984 & \textcolor{red}{\textbf{58.409}} \\
                          & CLIP-IQA $\uparrow$ & \textcolor{blue}{0.397} & 0.349 & 0.176 & 0.346 & 0.306 & \textcolor{red}{\textbf{0.475}} \\
                          & BRISQUE $\downarrow$ & 27.903 & \textcolor{blue}{20.427} & 51.052 & 25.936 & 45.095 & \textcolor{red}{\textbf{9.190}} \\
\multirow{-4}{*}{VideoLQ} & DOVER $\uparrow$ & 74.010 & 70.877 & 46.082 & \textcolor{blue}{74.450} & 72.853 & \textcolor{red}{\textbf{74.563}} \\ \hline
                          & MUSIQ $\uparrow$ & \textcolor{blue}{55.500} & 53.059 & 33.384 & 49.757 & 48.988 & \textcolor{red}{\textbf{58.096}} \\
                          & CLIP-IQA $\uparrow$ & \textcolor{blue}{0.538} & 0.500 & 0.386 & 0.495 & 0.411 & \textcolor{red}{\textbf{0.607}} \\
                          & BRISQUE $\downarrow$ & 33.255 & 36.247 & 56.907 & \textcolor{blue}{27.321} & 50.478 & \textcolor{red}{\textbf{11.725}} \\
\multirow{-4}{*}{AIGC29}  & DOVER $\uparrow$ & 60.308 & 78.846 & 59.582 & \textcolor{blue}{82.405} & 80.111 & \textcolor{red}{\textbf{83.547}} \\ \hline
\end{tabular}%
}
\end{table*}

\subsection{Implement Details.}
Stable Video Diffusion (SVD) is adopted as our foundational video diffusion model. The proposed method is trained on $8$ NVIDIA A100-80G GPUs with a batch size of $8$. 
We employ a constant learning rate of $8e-5$ with the AdamW optimizer.
More training details can be found in the appendix.

\subsubsection{Training and Testing Datasets.}
For training, we collect a large-scale high-resolution dataset from the website containing approximately $300k$ video clips.
To evaluate the performance of the proposed method, we utilize three types of testing datasets: the synthetic dataset, the real-world dataset, and the AIGC dataset.
For the synthetic datasets, we employ two publicly available datasets, UDM10 \cite{tao2017detail} and YouHQ40 \cite{zhou2024upscale}, for evaluation purposes. The real-world dataset consists of the publicly available VideoLQ \cite{chan2022investigating} dataset for testing.
Regarding the AIGC dataset, since no public dataset is available, we have collected a dataset containing $29$ videos (referred to as AIGC29) sourced from popular text-to-video generation models \cite{chen2023videocrafter1,chen2024videocrafter2,yang2024cogvideox}.

\subsubsection{Evaluation Metrics.}
We follow \cite{zhou2024upscale} and adopt different metrics to evaluate the frame quality and temporal coherency. For synthesis datasets, we employ PSNR, SSIM, LPIPS \cite{zhang2018unreasonable} and flow warping error $E_{warp}$ \cite{lai2018learning} for evaluation. In the case of the real-world and AIGC datasets, where ground-truth videos are unavailable, we adopt non-reference metrics including CLIP-IQA \cite{wang2023exploring}, MUSIQ \cite{ke2021musiq}, DOVER \cite{wu2023exploring}, and BRISQUE \cite{mittal2012no} for assessment.

\begin{figure}[t]
  \centering
  \includegraphics[width=0.98\linewidth]{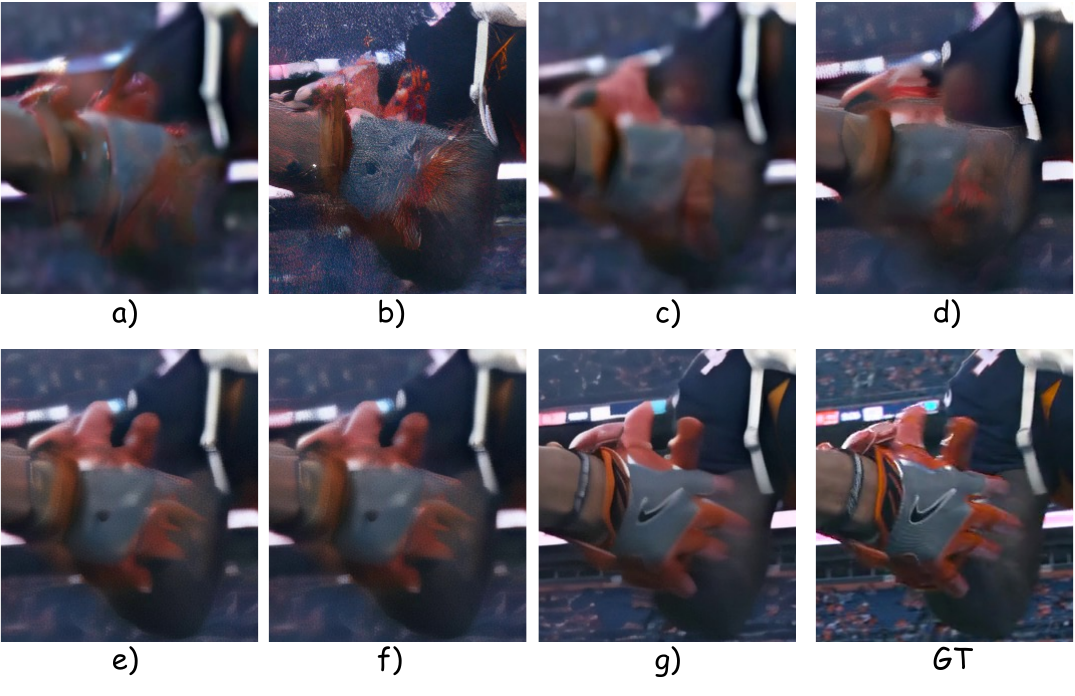}
  \caption{Qualitative ablation study of DAM-VSR.}
  \label{fig:ablation}
\end{figure}

\subsection{Ablation Study}

To validate the effectiveness of various components within DAM-VSR, we conduct an ablation study encompassing the following elements: Disentanglement Framework, Fine-tuning VAE Decoder and Motion-aligned Bidirectional Sampling. The experimental results are shown in Table \ref{tab:ablation}. The ``Baseline'' denotes that we do not employ any of the three components and only train a video ControlNet conditioned on low-quality videos. 
Additionally, we provide experimental results for video super-resolution conducted frame by frame using the ISR method, denoted as "Frame-by-frame SR." The corresponding qualitative results can be found in Fig. \ref{fig:ablation}.

\subsubsection{Appearance and Motion Decomposition Generation.}
We first investigate the impact of the proposed appearance and motion disentanglement framework for VSR. We train a video ControlNet conditioned on low-quality videos without appearance enhancement. During inference, we only utilize low-quality videos without reference image enhancement. This variant is denoted as ``w/o Disentanglement During Training and Inference". Furthermore, we utilize the same video ControlNet with ``w/o Disentanglement During Training and Inference'' but conduct reference image enhancement only during inference. The experimental results are denoted as ``w/o Disentanglement only During Training". 
Both of these experimental setups face a decrease in performance, which proves the effectiveness of our decomposition framework.

\subsubsection{Motion-aligned Bidirectional Sampling.} 
During inference, we predict the noise by including only a single disentangled forward generation at each step. We examine the effect of the motion-aligned bidirectional sampling, and the results are denoted as ``w/o Bidirectional Sampling" in Table \ref{tab:ablation}. 
The motion-aligned bidirectional sampling effectively enhances the fidelity of the generated video.

\subsubsection{Fine-tuning VAE Decoder.} 
The variant without fine-tuning the VAE decoder is also presented in Table \ref{tab:ablation} and is denoted as ``w/o FT-VAE-Decoder". 
Fine-tuning the VAE decoder significantly improves the LPIPS metric, decreasing from $0.382$ to $0.311$, demonstrating that fine-tuning the VAE is crucial for enhancing fidelity.

\subsection{Comparison with Other Methods}

To verify the effectiveness of our method, we compare it with several state-of-the-art VSR methods, including RealBasicVSR \cite{chan2022investigating}, Upscale-A-Video \cite{zhou2024upscale}, StableVSR \cite{rota2023enhancing}, MGLD-VSR \cite{yang2025motion} and VEnhancer \cite{he2024venhancer}.

\subsubsection{Quantitative Comparison.}
The quantitative comparison results on synthetic datasets, real-world datasets, and AIGC datasets are shown in Table \ref{tab:compare}.
For synthetic data, since fidelity is the most important factor, ResShift \cite{yue2024resshift} is adopted as the ISR method to improve the fidelity of the generated videos. For real-world data and AIGC data, due to the need to supplement more details and fix artifacts, we chose ISR methods with stronger generation capabilities, such as SupIR \cite{yu2024scaling} and InvSR \cite{yue2024arbitrary}. 
The proposed method achieves the best evaluation metrics in most of the synthetic datasets, including UDM10 \cite{tao2017detail} and YouHQ40 \cite{zhou2024upscale}. However, RealBasicVSR \cite{chan2022investigating} and MGLD-VSR \cite{yang2025motion} are trained on the REDS dataset, which prevents us from obtaining the best performance. Nevertheless, our method still achieves the second place. VEnhancer \cite{he2024venhancer} achieves the best $E_{warp}$ across all synthetic datasets, which can be attributed to its ability to provide super-resolution in the temporal dimension, effectively eliminating flickering. On real-world and AIGC datasets, the proposed method has achieved state-of-the-art performance across all evaluation metrics, significantly surpassing other methods in MUSIQ, CLIP-IQA, and DOVER by a substantial margin, demonstrating the high perceptual quality of the proposed method. Additionally, our method is compatible with multiple ISR approaches, allowing VSR performance improvement without retraining when a better ISR method is developed. The ability to select different ISR methods to meet various needs also demonstrates the flexibility of our disentanglement framework, which is not available in other methods.

\subsubsection{Qualitative Comparison.}
To demonstrate the visual effectiveness of the proposed method, we visualize the results on synthetic, real-world, and AIGC videos in Fig. \ref{fig:youtu}, Fig. \ref{fig:videolq}, and Fig. \ref{fig:aigc}. As shown in Fig. \ref{fig:youtu}, the proposed method effectively removes degradations and generates realistic details, showcasing high fidelity in the synthetic data. In real-world videos, as illustrated in Fig. \ref{fig:videolq}, our method recovers the most natural details than other methods, such as the texture on the spire of a building. On AIGC data, DAM-VSR effectively generates realistic fine details, such as clear armbands on astronauts' arms and distinct clouds in galaxies, as shown in Fig. \ref{fig:aigc}.

\section{Limitation}

The first limitation is the computational process associated with bidirectional sampling, which can consume additional time during sampling. The second limitation pertains to the influence of the reference ISR results on the overall quality of video generation. 
The realistic details derived from ISR can be propagated to the remaining frames through the temporal layer. When ISR does not perform well, it may fail to generate satisfactory VSR results. However, the development of ISR is currently ahead of that of VSR. ISR outperforms VSR in both visual effects and evaluation metrics. Thus, our framework is viable until VSR advances beyond ISR.

\section{Conclusion}

In this paper, we propose DAM-VSR, an appearance and motion disentanglement framework for real-world VSR. The proposed method disentangles VSR into appearance enhancement and motion control tasks, benefiting from the ISR model. Furthermore, to handle longer input videos, we propose a motion-aligned bidirectional sampling strategy consisting of a disentangled forward generation and a disentangled backward generation. Quantitative and qualitative comparisons of synthetic data, real-world data, and AIGC data demonstrate superior performance in generating realistic details.





\begin{acks}
This work was supported in part by the National Natural Science Foundation of China (Grant No. 62372480, No. 62406089), in part by the Guangdong Basic and Applied Basic Research Foundation (No. 2025A1515010782, No. 2023A1515012839, No. 2025A1515011361), in part by Huawei Gift Fund (No. HUAWEI25IS02), and in part by HKUST-MetaX Joint Lab Fund (No. METAX24EG01-D).
\end{acks} 

\bibliographystyle{ACM-Reference-Format}
\bibliography{sample-bibliography}

\appendix

\begin{figure*}[htb]
  \centering
  \includegraphics[width=0.98\linewidth]{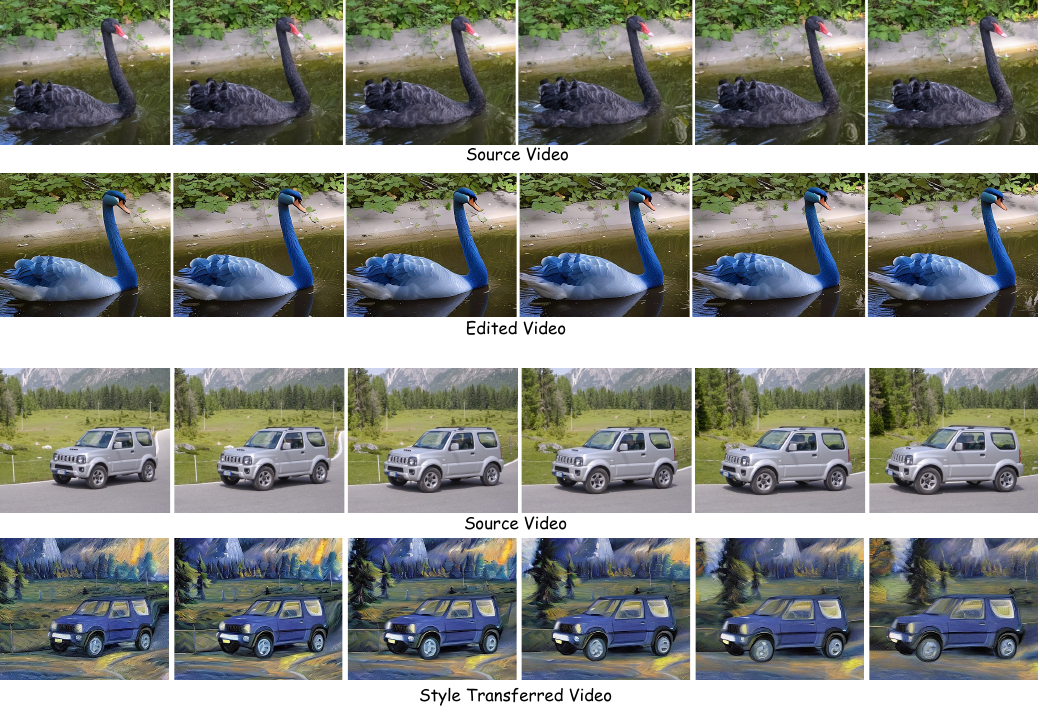}
  \caption{The experimental results of the proposed framework for other applications, including video editing and video style transfer.}
  \label{fig:application}
\end{figure*}

\begin{figure*}[htb]
  \centering
  \includegraphics[width=0.98\linewidth]{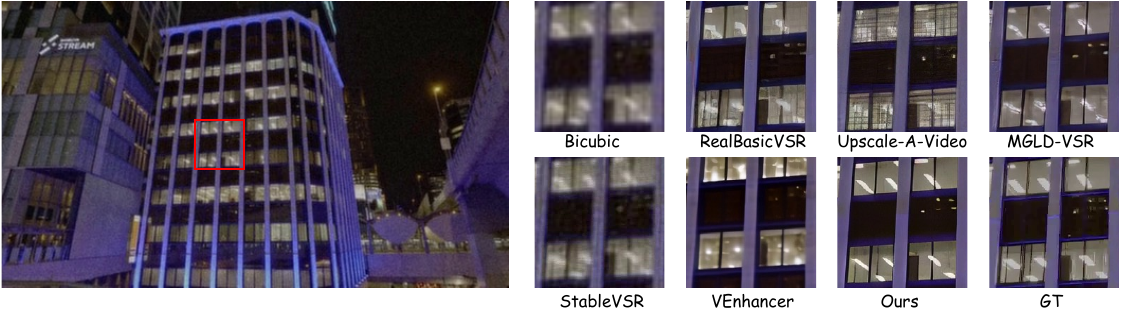}
  \caption{Qualitative comparison on synthetic low-quality videos from YouHQ40.}
  \label{fig:youtu}
\end{figure*}

\begin{figure*}[htb]
  \centering
  \includegraphics[width=0.98\linewidth]{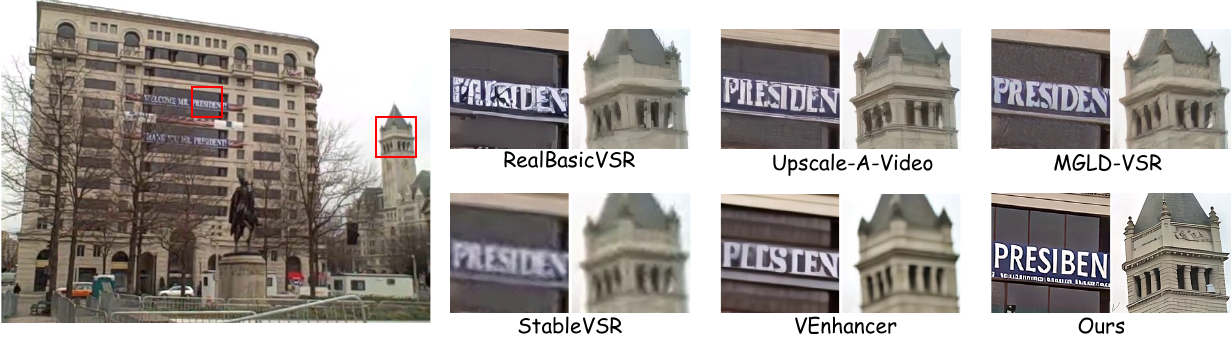}
  \caption{Qualitative comparison on real-world videos from VideoLQ.}
  \label{fig:videolq}
\end{figure*}

\begin{figure*}[htb]
  \centering
  \includegraphics[width=0.98\linewidth]{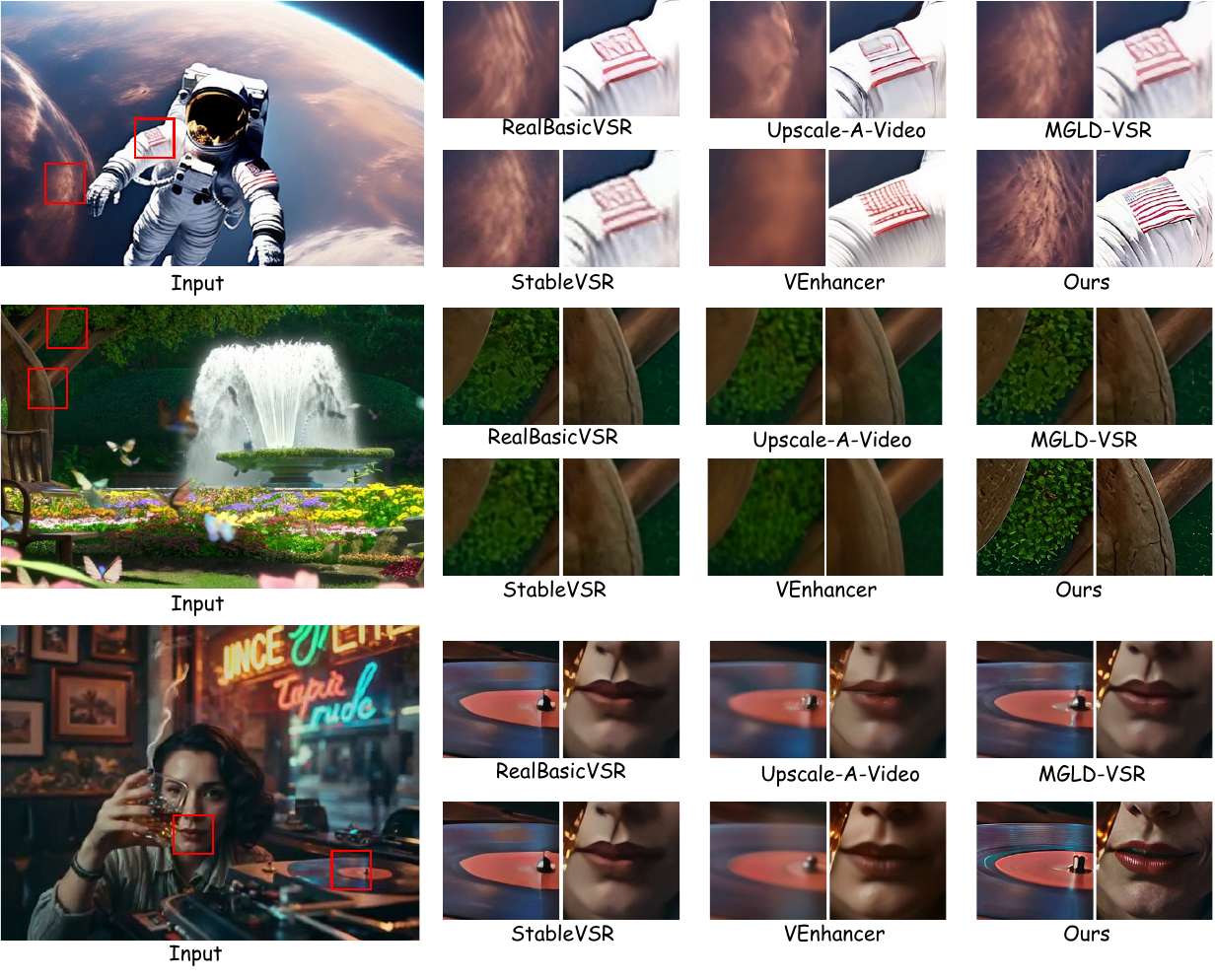}
  \caption{Qualitative comparison on AIGC videos.}
  \label{fig:aigc}
\end{figure*}